\documentclass[letterpaper]{article} 
\usepackage{aaai25}  
\usepackage{times}  
\usepackage{helvet}  
\usepackage{courier}  
\usepackage[hyphens]{url}  
\usepackage{graphicx} 
\usepackage{natbib}  
\usepackage{caption} 
\usepackage{algorithm}
\usepackage[noend]{algpseudocode}
\usepackage{newfloat}
\usepackage{listings}
\usepackage{placeins}
\DeclareCaptionStyle{ruled}{labelfont=normalfont,labelsep=colon,strut=off} 
\floatstyle{ruled}
\newfloat{listing}{tb}{lst}{}
\floatname{listing}{Listing}
\title{Improving Controller Generalization with Dimensionless Markov Decision Processes}
\author{
		Valentin Charvet \\
		Sebastian Stein \\
		Roderick Murray-Smith \\
}
\affiliations{
	School of Computing Science, University of Glasgow

}

\usepackage{tikz}
\usetikzlibrary{shapes.geometric, arrows, fit}
\tikzstyle{hidden_node} = [circle, text centered, draw=black, fill=gray!30]
\tikzstyle{observed_node} = [circle, text centered, draw=black]
\tikzstyle{box} = [rectangle, text centered, draw=black]
\tikzstyle{arrow} = [thick, ->, >=stealth, align=center]
\tikzstyle{dashed_arrow} = [dashed, ->, >=stealth, align=center]

\def\vtheta{{\bm{\theta}}}

\def\vc{{\bm{c}}}
\def\vs{{\bm{s}}}
\def\vx{{\bm{x}}}
\def\vy{{\bm{y}}}
\def\va{{\bm{a}}}
\def\gC{{\mathcal{C}}}
\def\gS{{\mathcal{S}}}
\def\gA{{\mathcal{A}}}
\def\gR{{\mathcal{R}}}
\def\gM{{\mathcal{M}}}
\def\sR{{\mathbb{R}}}
\newcommand{\E}{\mathbb{E}}
\def\gO{{\mathcal{O}}}
\def\mC{{\bm{C}}}

\usepackage{amsthm, amsmath, amsfonts, bm, dsfont}
\newtheorem{remark}{Remark}
\newtheorem{defn}{Definition}

\DeclareMathOperator*{\argmax}{arg\,max}

\usepackage{subcaption}
\usepackage{booktabs}

\begin{document}

\maketitle

\begin{abstract}
Controllers trained with Reinforcement Learning tend to be very specialized and thus generalize poorly when their testing environment differs from their training one.
We propose a Model-Based approach to increase generalization where both world model and policy are trained in a dimensionless state-action space.
To do so, we introduce the Dimensionless Markov Decision Process ($\Pi$-MDP): an extension of Contextual-MDPs in which state and action spaces are non-dimensionalized with the Buckingham-$\Pi$ theorem.
This procedure induces policies that are equivariant with respect to changes in the context of the underlying dynamics.
We provide a generic framework for this approach and apply it to a model-based policy search algorithm using Gaussian Process models.
We demonstrate the applicability of our method on simulated actuated pendulum and cartpole systems, where policies trained on a single environment are robust to shifts in the distribution of the context.
\end{abstract}

%
 \begin{links}
     \link{Code}{https:/****}
\end{links}

\section{Introduction}
One of the main obstacles for deploying controllers trained with Reinforcement Learning (RL) in the real world is their lack of resilience to perturbations and noise that are absent during training.
This problem of distribution shift has mostly been investigated in the supervised and unsupervised learning settings. 
Though the question can be phrased similarly in sequential decision-making, solving it remains difficult because of the dynamic nature of RL.
Firstly, because errors and approximations accumulate during planning and rollout, and secondly because the closed-loop nature of the learning process incurs a loss of identifiability \cite{Ljung1989}.
The issue is even more prevalent in Offline RL because of the lack of training data in some regions of the state-action space and the impossibility to collect more.

In this paper, we focus our work on perturbations that affect the environment dynamics only.
On the other hand, we assume the reward function is known and remains the same throughout the experiments.
The perturbations of the underlying transition kernel cause non-stationarity dynamics.
These can be caused by hardware wear-and-tear, feedback loops or external perturbations and is admitted to be one of the main challenges to be solved for deploying RL agents in the real world \cite{dulac2021challenges}.


\begin{figure}[ht]
\begin{subfigure}{.5\textwidth}
\centering
\begin{tikzpicture}[node distance=.4\textwidth, scale=0.5]
\node (agent) [box] {Agent};
\node (env) [box, left of=agent] {Environment};

\draw [arrow] (env.north) to[bend left] node[anchor=south] {$s_t$} (agent.north);
\draw [arrow] (agent.south) to[bend left] node[anchor=north] {$a_t$} (env.south);
\end{tikzpicture}
\caption{Markov Decision Process.}
\end{subfigure}
\begin{subfigure}{.5\textwidth}
\centering
\begin{tikzpicture}[node distance=.4\textwidth, scale=0.5]
\node (agent) [box] {Agent};
\node (env) [box, left of=agent] {Environment};
\node (context) [box, above of=env, yshift=-2.5cm] {Context};
\draw [arrow] (env.north) to[bend left] node[anchor=south] {$s_t$} (agent.north);
\draw [arrow] (agent.south) to[bend left] node[anchor=north] {$a_t$} (env.south);
\draw [arrow] (context.west) to[bend right] (env.west);
\end{tikzpicture}
\caption{Contextual MDP}
\end{subfigure}
\caption[Markov Decision Processes]{High level view of an agent interacting with its environment.}
\label{tikz:MDP_high_level}
\end{figure}
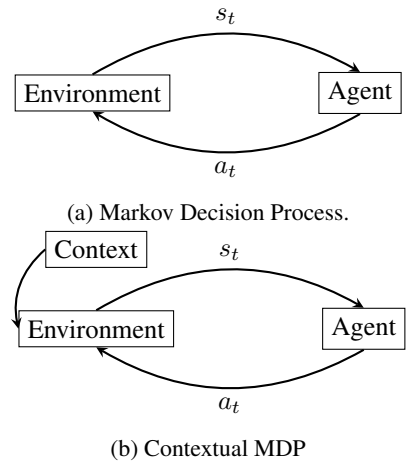

The method we propose sheds a new light on a generalization method based on \emph{Augmented World Models} \cite{pmlr-v139-ball21a}. 
In this work, the authors propose to increase the zero-shot generalization of a control policy learned offline from a single environment.
To do so, they rescale the observations by a factor inferred from data.
Our work proposes a similar transformation that is instead inferred from the dimensions of the variables using the Buckingham theorem \cite{buckingham1914physically}.
This theorem has also been applied to transfer learning problems for system identification \cite{electronics13112041} and control \cite{math12050709} in robotics.

In this work, we propose a transformation of the state-action spaces of Contextual-MDPs using a non-dimensionalising power law informed by the physics of the environment. 
This transformation, providing the context is observable, allows controllers trained in a single nominal environment to maintain optimal performance in the presence of a distribution shift. 
We present the empirical benefits of this approach in section \ref{sec:experiments} on the pendulum and cartpole environments.

\section{Previous Work}
\subsection*{Robust Reinforcement Learning}
Robustness can be achieved by optimizing a \emph{pessimistic objective}.
This is often referred to as the \emph{Robust Markov Decision Process} (MDP) framework \cite{doi:10.1287/moor.1120.0566/, eysenbach2021maximum}, which can be solved by approximate dynamics programming \cite{mankowitz2018learning, pmlr-v32-tamar14} or within Maximum a Posteriori Policy Optimization \cite{mankowitz2019robust}.
Such methods go back to 2005 \cite{morimoto2005robust} where the authors apply an actor-critic algorithm where the controller attempts to correct for disturbances generated by an internal agent.
More recently \cite{pinto2017robust} apply a similar method with neural networks.
In essence, these methods solve a minimax optimization problem to account for worst-case scenarios.
\cite{pmlr-v115-derman20a} defines an Uncertainty-Robust Bellman Equation and derives a robust TD error from it. This general framework was empirically verified in both discrete and high-dimensional continuous domains.
Other types of methods inject noise in the policy or the model in order to prevent overfitting \cite{charvet2021, igl2019generalization}.
These optimization procedures tend to yield controllers that are overly conservative, as they generalize quite well at the cost of losing optimality even on IID data.

Some other meta-learning approaches rely on \emph{domain randomization}.
These consist in training in multiple version of the environment (i.e. several contexts) so as to disentangle local and global properties of the task \cite{saemundsson2018meta, kupcsik2013data, charvet2021}.
All of these approaches however require access to a white-box simulator, on which we can intervene to change its properties. 

On the other hand, augmenting the set of initial hypotheses may increase the model and policy ability to learn and generalize with no additional data \cite{van2021multi, NEURIPS2022_29e4b51d}.
Successes on zero-shot transfer have been increased with causal models \cite{pmlr-v70-kansky17a,pmlr-v229-huang23c}.
There are also recent works that have studied the generalization problem but in the visual domains \cite{NEURIPS2023_43b77cef, NEURIPS2023_6692e1b0}

The issue of distribution shift is also a concern for Offline RL \cite{Levine2020}. In that specific setting however, it is not caused by non-stationarity but by the lack of training data in regions the offline-optimal policy visits. Several model-based methods propose to bypass it by means of regularization. MOReL and variants \cite{NEURIPS2020_f7efa4f8, pmlr-v202-kim23q} construct a pessimistic MDP and use a mechanism to detect unknown state-actions in order to split the space between regions of low and high uncertainty. MOPO \cite{NEURIPS2020_a322852c} also optimizes the policy in a surrogate MDP, where the reward is penalized by the model error. Both maximize a lower bound of the true objective. While both methods are conceptually similar, MOPO resorts to a softer penalty than MOReL.
Other methods rely on Importance-Sampling schemes such as \cite{NEURIPS2023_ae8b0b58, NEURIPS2021_949694a5, NEURIPS2023_0ff3502b}

Like \cite{pmlr-v115-derman20a}, we believe Bayesian models are well-suited for the generalization task in RL.
In the domain of classical methods, \emph{Dual Control} \cite{882466} maintains a probabilistic estimation of the plant parameters to derive robust adaptive controllers.
This is due to the way a Bayesian can reason about an infinite number of models by means of a distribution, and integrate over all the possibilities, weighted by how likely they are.
In contrast, worst-case approaches only consider a subset of models that include the most pessimistic realizations.

\subsection*{Buckingham-$\Pi$ theorem} 
The Buckingham theorem was first introduced in \cite{buckingham1914physically} as a way to reduce the number of parameters to control for collecting experimental data in physics.
It was instrumental in the interpretation of fundamental quantities such as the Reynolds number in fluid dynamics \cite{lee2021dimensional}.
A consequence of the theorem is a method for transforming the input variables into dimensionless quantities, as we further explain in section \ref{sec:methods}.
More importantly \cite{43c9340b-02e2-39cf-8358-16a6c89c707f, doi:10.1080/00401706.2017.1291451} demonstrate that dimensionless variables are maximal invariant statistics with respect to scale transformation in fundamental dimensions.
More recent work in the machine learning literature has demonstrated how estimators trained on dimensionless variables are able to generalize predictions outside the support of training data \cite{OPPENHEIMER2023111810, kumar2018machine, villar2022dimensionless, math12050709}.

\section{Background}
\subsection{Contextual Markov Decision Process}
The evaluation of controllers trained with RL is often done in the same environment they have been trained on.
In such cases, the unique training and evaluation metric is the return:
\begin{equation}
    R(\pi, f) = \E \left[ \sum_{t=0}^T \gamma^t r_t \bigg| a_t \sim \pi(s_t), s_{t+1} \sim f(s_t, a_t)\right].
\label{eq:return}
\end{equation}
While this provides a good test-bed for designing and comparing algorithms, it tends to oversimplify what would actually happen in the real world, where dynamics can be non-stationary \cite{dulac2021challenges}.
Physical wear-and-tear or hidden feedback loops \cite{NIPS2015_86df7dcf} can cause a significant distribution shift which hinders the ability of a controller to stabilize the system at its equilibrium.
Though it is not the only approach to illustrate this drift, we will assume the dynamics of the MDP are subjected to a set of hidden variables that impact its one-step transitions.
We follow the notations from \cite{kirk2022survey} and call this set of variables the \emph{context}.

From this follows the definition of \emph{Contextual Markov Decision Process (C-MDP)} \cite{hallak2015contextual, doshi2016hidden, NEURIPS2021_d5ff1353}, characterized by the following transition kernel
\begin{equation}
s_{t+1} \sim f_{|\vc}(s_{t+1} | s_t, a_t; \vc).
\label{eq:context_markov_kernel}
\end{equation}
The difference with usual MDPs is illustrated in figure \ref{tikz:MDP_high_level}, where the context acts as a set of confounding variables on the dynamics.

\begin{remark}
C-MDPs can alternatively be viewed as a Partially Observed MDP with an emission function that constantly returns the observed state $\gO(s_t, \vc) = s$.
They are also in close connection with Latent MDPs \cite{NEURIPS2021_cd755a6c} where the context is sampled at random at the beginning of each episode.
\end{remark}

Because we consider that the context is slowly evolving, we assume in all the following analysis that it is sampled from an unknown distribution $p(c)$ at the beginning of an episode and remains static throughout its duration.
The control objective in this setting can then be extended as
\begin{equation}
\underset{\pi}{\max} \left\{ \underset{\vc \sim p(\vc)}{\E}  \left[ R(\pi, f_{|\vc}) \right] \right\}.
\label{eq:objective_CMDP}
\end{equation}
In a similar way as in supervised learning, we can define the \emph{Generalization Gap} as the discrepancy between returns obtained in the training environment and the testing one,
\begin{equation}
  \text{GenGap}\left(\pi, \vc_{train}, \vc_{test}\right) = R\left(\pi, f_{|\vc_{train}}\right) - R\left(\pi, f_{|\vc_{test}}\right).
\label{eq:GenGap}
\end{equation}
A robust controller will be able to achieve a low generalization gap for a wide set of testing contexts.
This may, however, come at the cost of being overly conservative, meaning the controller will not be optimal even in the training environment.
Trading-off optimal performance and robustness is at the core of robust RL research and our approach for doing so consists in increasing the set of hypotheses rather than collecting additional training data. 

\subsection{Dimensionless Machine Learning}
\subsubsection*{Units-Typed Spaces}
Before jumping into the details of the Buckingham-$\Pi$ theorem \cite{buckingham1914physically}, we need to explain what a physical measurement and dimension are, since they are often ignored in the machine learning practice.
Following the bracket notation from \cite{sonin2001dimensional} where $[X]$ denotes the dimension of variable $X$ and $\{X\}$ its magnitude, a physical measurement may be written as
\begin{equation}
 X = \left\{X\right\} \left[X\right].
\label{eq:magnitude-dimension}
\end{equation}
In mechanics for instance, every measurement can be expressed with the elementary dimensions of time $T$, length $L$ and mass $M$.
The measure of a distance for example will have the dimension of a length $\left[L\right]$, and acceleration a length per time squared $\left[L T^{-2}\right]$.
\begin{itemize}
  \item Two quantities  $X$ and $Y$ can be added provided $\left[X\right]=\left[Y\right]$ and the resulting quantity has magnitude $\{X+Y\} = \{X\}+\{Y\}$ and dimension $[X+Y]=[X]=[Y]$.
\item Two quantities can be multiplied whatever their dimensions are and $\{X\times Y\}=\{X\}\times \{Y\}$, $[X\times Y]=[X]\times [Y]$.
\item A quantity can be raised to the power of a rational fraction $\gamma \in \mathbb{Q}$ with
$X^\gamma = \{X\}^\gamma [X]^\gamma$.
\end{itemize}

\subsubsection*{Buckingham-$\Pi$ theorem}
In essence, the Buckingham-$\Pi$ theorem states that if a physical system is described as a function of $d$ independent variables with $k$ elementary dimensions, then it can be equivalently described by $d-k$ dimensionless variables.
For instance, a system described by an equation
\begin{equation}
    f(x_1, \dots, x_1) = 0,
\label{eq:dimensional_equation}
\end{equation}
is equivalent to 
\begin{equation}
    f_{\Pi} (\Pi_1, \dots, \Pi_{d-k}) = 0.
\label{eq:dimensionless_equation}
\end{equation}
The variables in equation (\ref{eq:dimensionless_equation}) are called \emph{$\Pi$-groups} and are obtained through a power-law of the dimensional variables such as,
\begin{equation}
    \Pi_j = \prod_{i=1}^d x_i^{z_{i,j}}, z_{i,j} \in \mathbb{Z}.
\label{eq:power_law}
\end{equation}
Each $\Pi$-group is dimensionless, meaning $\forall j \in (1, \dots, d-k)~ [\Pi_j] = 1$.
The $z_{i,j}$ coefficients are, in general, not unique and found by solving a system of diophantine equations.

\subsubsection*{Dimensionless Cartpole}
As an illustrative example, we now demonstrate in practice how a cartpole system can be non-dimensionalised using this method.
We provide the details of the derivations in the appendix \ref{app:cartpole_pi_groups} and \ref{app:pendulum_pi_groups} for the cartpole and pendulum respectively.
This system can be described by a C-MDP with state space $\mathcal{S}=(x, \theta, \dot{x}, \dot{\theta})$ and action space $\mathcal{A}=(u)$, described in table \ref{tab:cartpole_state_variables}.
The context consists of the variables $\mathcal{C}=(L, M, g)$, the length and mass of the pendulum as well as the gravity field. 
We omit the friction coefficients since a shift in them has little impact on the dynamics.

\begin{table}[ht]
\begin{center}
  \begin{tabular}{|c|c|c|c|}
    \toprule
    variable & natural & dimension & $\Pi-group$ \\
    \midrule[.5pt]
    cart position & $x$ & $L$ & $\frac{x}{L}$ \\
    angle & $\theta$ & $1$ & $\theta$ \\
    cart speed & $\dot{x}$ & $L.T^{-1}$ & $\frac{\dot{x}}{\sqrt{Lg}}$ \\
    angular speed & $\dot{\theta}$ & $T^{-1}$ & $\dot{\theta}\sqrt{\frac{g}{L}}$ \\
    control force & $u$ & $M.L.T^{-2}$ & $\frac{u}{Mg}$ \\
    \bottomrule
  \end{tabular}
\end{center}
\caption{Natural and dimensionless state-action variables}
\label{tab:cartpole_state_variables}
\end{table}

The dimensionless variables in table \ref{tab:cartpole_state_variables} are obtained by multiplying each natural variable by factors of the elements of the context vector. 
Note that it can be done provided the units of the vector of dimensions of the context is of full rank. 
We also emphasize that angles are naturally dimensionless quantities since they are a quotient of two lengths. 
In practice, we parameterize the angle with sine and cosine functions.

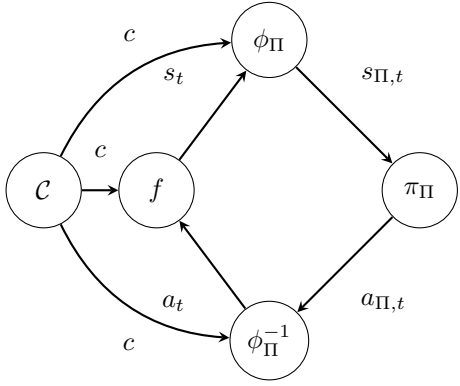
\begin{figure}[htb]
	\centering
	\begin{tikzpicture}[node distance=1.5cm, scale=1cm, minimum width=1cm, minimum height=1cm]

		\node(env)[observed_node]{$f$};
		\node(phi_pi)[observed_node, right of=env, yshift=2cm]{$\phi_\Pi$};
		\node(policy)[observed_node, right of=env, xshift=2cm]{$\pi_\Pi$};
		\node(phi_inverse)[observed_node, right of=env, yshift=-2cm]{$\phi_\Pi^{-1}$};
		\node(context)[observed_node, left of=env]{$\gC$};

		\draw[arrow](env) --node[anchor=south east]{$s_t$} (phi_pi);
		\draw[arrow](phi_pi) --node[anchor=south west]{$s_{\Pi,t}$}(policy);
		\draw[arrow](policy) --node[anchor=north west]{$a_{\Pi,t}$} (phi_inverse);
		\draw[arrow](phi_inverse) --node[anchor=north east]{$a_t$}(env);

		\draw[arrow](context) -- node[anchor=south]{$c$}(env);
		\draw[arrow](context) to[bend left] node[anchor=south]{$c$}(phi_pi);
		\draw[arrow](context) to[bend right] node[anchor=north]{$c$}(phi_inverse);

	\end{tikzpicture}
	\caption{Interaction within a $\Pi$-MDP}
	\label{tikz:Pi-MDP}
\end{figure}

\section{Methods}
\label{sec:methods}
\subsection{Dimensionless MDP}
We call $\Phi_\Pi$ the function that transforms the natural variables into the $\Pi$-groups and $\Phi_\Pi^{-1}$ the inverse operation.
This transformation depends on the current value of the context.
In order to reason about control policies within a dimensionless state-space, we introduce a new concept we call the $\Pi$-MDP.
The $\Pi$-MDP is a generalization of the C-MDP, which is equipped with a dimensionless invertible transformation $\Phi_\Pi$ that transforms the state and actions spaces depending on the context vector $\vc$.
\begin{defn}[$\Pi$-MDP] 
The dimensionless Markov Decision Process or $\Pi$-MDP is a MDP in which the state and action spaces are dimensionless.
They can be written
\begin{equation}
  \gM_\Pi = \left(\gS_\Pi, \gA_\Pi, \gR, f_\Pi \right),
\end{equation}
where $\gR$ is the reward function and $f_\Pi$ the transition kernel that takes values in the dimensionless sate-action space is defined as, $f_\Pi = f \circ \Phi_\Pi$,
where $\circ$ denotes the functional composition and $\Phi_\Pi$ is the non-dimensionalization transformation.
\end{defn}
Within a step of an episode, a state is generated by the contextual Markov kernel given the previous state, action and context such as
$\vs_{t+1} \sim f(\vs_t, \va_t; \vc)$.
This state is then non-dimensionalized with $\vs_{\Pi, t+1}=\Phi_\Pi (\vs_{t+1})$ such that the policy takes a decision in a dimensionless space. 
This action is then transformed back into natural space $a_{t+1} = \Phi_\Pi^{-1}(a_{\Pi, t+1})$. 
This process is repeated until the end of the current episode and is summarized on figure \ref{tikz:Pi-MDP}.
Throughout the experiments, we assume the context $\vc$ is observable and remains static throughout the duration of an episode.
The pseudo-code for interacting within a $\Pi$-MDP is given in algorithm \ref{alg:interaction-Pi-MDP}.

\begin{algorithm}[H]
\caption{Interaction in a $\Pi$-MDP}
\begin{algorithmic}[1]
\State \textbf{Input} policy $\pi_\Pi$, dimensionless feature map $\Phi$, initial state $s_0$
\State $s_t \gets s_0$
\For{$t=1, \dots, T$} \Comment{number of steps of an episode}
  \State $s_{\Pi, t} =  \Phi(s_t)$   \Comment{non-dimensionalize observation}
  \State $a_{\Pi,t} =  \pi_\Pi(s_{\Pi,t})$ \Comment{choose dimensionless action}
  \State $a_t = \Phi^{-1}(a_{\Pi,t})$   \Comment{dimensionalize action}
  \State $s_t \gets f(s_t ,a_t)$ \Comment{1-step Markov transition}
  \State $r_t = R(s_t)$
\EndFor
\Return{$\sum r_t$}   \Comment{Cumulative Rewards}
\end{algorithmic}
\label{alg:interaction-Pi-MDP}
\end{algorithm}

\subsection{Model-Based Reinforcement Learning}
Model-Based Reinforcement Learning (MBRL) is a class of RL algorithms in which the policy is trained on data generated by a \emph{world model}.
For this reason, such algorithms are often called \emph{indirect methods} as opposed to model-free approaches that optimize their decisions using data directly collected in the environment.

The first requirement for such MBRL algorithms is the dynamics model, an estimator that mimics the behaviour of the MDP transition kernel,
\begin{equation}
\hat{f}: (\vs_t,\va_t) \mapsto \hat{\vs}_{t+1}.
\label{eq:transition_model}
\end{equation}
This model is subsequently trained to predict one-step transitions using the batches of data collected so far.
It is therefore a multidimensional regression problem where the inputs are the state-action vectors $\tilde{\vx}=(\vs, \va)\in \sR^{d+f}$ and the targets are the successor states $\vy =(\vs_{t+1}-\vs_t) \in \sR^d$.
Because the target $\vy$ are vectors, MBRL methods are more sample-efficient than model-free methods since the latter learn from scalar reward signals instead.
The model can then be queried to generate one-step transitions or whole trajectories with a parametric policy $\pi_\vtheta$.
We write the closed-loop dynamics as
\begin{equation}
f_\vtheta: s \mapsto f(\vs' | \vs, \pi_\vtheta(s)).
\label{eq:closed-loop_dynamic}
\end{equation}
 Its estimate counterpart $\hat{f}_\vtheta$ is able to generate whole trajectories by functional composition in order to predict the future state of a system under the current policy.
To do so, we start from an initial state $\vs_0$ and iterate the predictions until desired time.
\begin{equation}
\hat{s_t} = \underbrace{\hat{f}_\vtheta \circ \cdots \circ \hat{f}_\vtheta(\vs_0)}_{t~ \text{times}}.
\label{eq:long_term_prediction}
\end{equation}

This ability to query the model to predict long-term states of the system is what makes this type of method useful.
It can generate trajectories $\tau=(\vs_0, \cdots, \vs_t)$ of arbitrary size.
Assuming we know the reward function $r$, we can compute the simulated expected sum of rewards from the future state predictions.
The policy search objective can therefore be written as,
\begin{equation}
    \hat{R}(\vtheta) = \underset{\hat{f}^\vtheta}{\mathbb{E}} \left[ \sum_{t=0}^T r(\hat{\vs_t}) | s_0 \right] .
\label{eq:return_MB}
\end{equation}
This quantity serves as a proxy for the return that would be obtained by rolling out in the environment.
The objective (\ref{eq:return_MB}) is very similar to (\ref{eq:return}) but with the expectation measured by the approximate dynamics.
A controller is optimal for the model if it maximizes that quantity (\ref{eq:return_MB}), however there is no guarantee that $\argmax_\vtheta \hat{R} = \argmax_\vtheta R$ because of \emph{model bias}.
Because during training the policy is only exposed to data generated by the model, any discrepancy with the true dynamics will reflect on the quality of the policy.
Moreover, due to compounding errors in equation (\ref{eq:long_term_prediction}), estimating the future states is a difficult task.
One solution is to use the model on short rollouts only \cite{NEURIPS2019_5faf461e}. 
Alternatively, a probabilistic model is able to eliminate most of the bias associated with predictions.
Given a state-action input, a probabilistic model will predict a distribution over plausible future states.
Hence, rolling it out with (\ref{eq:long_term_prediction}) yields a distribution of trajectories ${p}(\tau)$.
If the model is wrong, the trajectories will be associated with high levels of uncertainty that will propagate to the estimation of (\ref{eq:return_MB}).

To optimize the parameters of the policy, different algorithms use different gradient of return estimation schemes like reparameterization trick \cite{kingma2013auto, pmlr-v89-xu19a} or likelihood ratio \cite{Williams1992} to backpropagate derivatives through sampling the model \cite{JMLR:v21:19-346}.
Suppose $\hat{\nabla}_\vtheta \hat{R}(\vtheta)$ is an unbiased estimation of the gradient, we can optimize the policy with stochastic steps in the ascending direction
\begin{equation}
\vtheta \leftarrow \vtheta + \eta \hat{\nabla}_\vtheta \hat{R}(\vtheta),
\label{eq:gradient_ascent_step}
\end{equation}
with $\eta$ the learning rate.
Between each episode, the policy is optimized with the current dynamic model until the expected return plateaus.
Then, the policy collects a new episode of data which is fed into training the model.
The model will improve using the new data, and so on, until some measure of convergence is reached.

We extend the subclass of model-based policy gradient methods with Gaussian Process priors \cite{Deisenroth2011, pmlr-v80-parmas18a, Romeres2022dec, cowen2022samba} because their ability to estimate uncertainty eliminates most of the bias.
This ability to plan with uncertainty has allowed model-based algorithms to compete with their model-free alternatives \cite{schrittwieser2020mastering, NEURIPS2018_3de568f8, NEURIPS2019_5faf461e}.
Instead of optimizing the controller in the natural state-space view, we do it in its dimensionless counterpart.
This transformation essentially renders the controller equivariant to context changes and so is able to generalize outside its training support.

\subsection{Dimensionless Policy Search}
We design a control policy \ref{eq:dimless_policy} that acts in a dimensionless space, making it equivariant to changes in the context vector.
\begin{equation}
  \pi_\Pi(s, \vtheta) = \pi_\Pi(\Phi_\Pi(s); \vtheta)
\label{eq:dimless_policy}
\end{equation}

We introduce a new algorithm $\Pi$-PILCO: \emph{Dimensionless Probabilistic Inference for Learning COntrol}, a variation of the data efficient PILCO algorithm that performs policy search within a dimensionless state space.
Let us note that the methodology can in principle, be applied to any MBRL algorithm provided the state and action space can be non-dimensionalized with the Buckingham-$\Pi$ theorem.

In essence, the algorithm is not very different from the one in natural space.
The difference here is that both dynamics model and policy have dimensionless inputs and outputs.
When the policy is interacting with the MDP, it non-dimensionalizes the observations, returns a dimensionless control, which is then projected back in natural space before being sent to environment.
The procedure is described in extensive detail in algorithm \ref{alg:interaction-Pi-MDP}.

\begin{algorithm}[H]
\caption{Dimensionless Policy Search - $\Pi$-PILCO}
\begin{algorithmic}[1]
\State \textbf{Input} policy $\pi_{\Pi, \vtheta}$, dimensionless feature map $\Phi$, dimensionless model $\hat{f}_\Pi$
\For{$i = 1, \dots, P$} \Comment{number of epochs}
  \State $s_t \sim \rho_0$ \Comment{sample initial state}
  \State $s_{\Pi, t} =  \Phi(s_t)$
  \State $R = 0$
  \For{$t = 1, \dots, H$} \Comment{prediction horizon}
    \State $a_{\Pi,t} = \pi_\Pi(s_{\Pi,t}; \vtheta)$
    \State $s_{\Pi,t} \gets \hat{f}_\Pi(s_{\Pi,t}, a_{\Pi,t})$
    \State $R \gets R + R(s_{\Pi, t})$
  \EndFor
  \State $\vtheta \gets \vtheta + \nabla_\vtheta R$ \Comment{gradient step}
\EndFor
\Return{$\pi_{\Pi, \vtheta}$}
\end{algorithmic}
\label{alg:PS-Pi-MDP}
\end{algorithm}
The optimization objective is the same as previously described in equation (\ref{eq:return_MB}), the difference lies in the way trajectories are computed.
In equation (\ref{eq:long_term_prediction}), the closed-loop dynamics iterate one-step predictions on the natural state-action spaces.
So at each time step, the policy selects a dimensionless action based on dimensionless observations and the model predicts the next state (or distribution thereof).
Additionally, we compute the reward at each step by applying the inverse Buckingham transformation to the dimensionless state.
We repeat the procedure until a horizon $H$ is reached and the local rewards are summed to estimate the gradient of the return.
For ease of exposition, algorithm \ref{alg:PS-Pi-MDP} details the policy search methodology based on the Reparameterization Trick as in \cite{pmlr-v80-parmas18a}.

\begin{figure}[htb]
\centering
\includegraphics[width=0.5\textwidth]{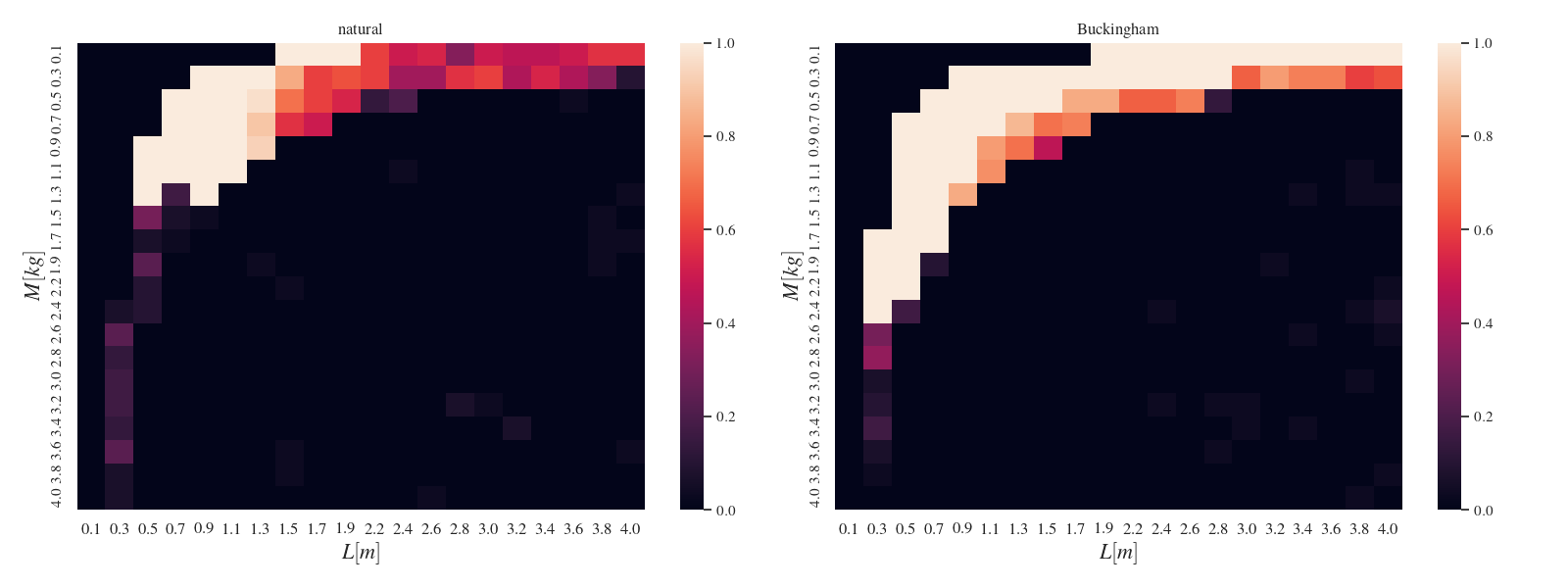}
\caption[Pendulum 2D variation of $M$ and $L$]{Pendulum success rates on the pole length when both $M$ and $L$ are varying for the natural (left) and dimensionless (right) controllers. Brighter values indicate higher success rates.}
\label{fig:pendulum-2D_ML-returns}
\end{figure}

\section{Experiments}
\label{sec:experiments}
\begin{table}[ht!]
\begin{center}
  \begin{tabular}{|c|c|c|c|}
    \toprule
    Environment & $L[m]$ & $M[kg]$ & $g[m.s^{-2}]$ \\
    \midrule[.5pt]
    Pendulum & 1 & 1 & 10 \\
    Cartpole &1 & 0.1 & 9.81\\
    \bottomrule
  \end{tabular}
\end{center}
\caption[Nominal context value]{Nominal context value for the cartpole and pendulum environments}
\label{tab:nominal_context_values}
\end{table}
We will evaluate our algorithm on two second-order systems, the first is the underactuated pendulum.
The cartpole is a slightly more complicated one, where a pendulum is attached to a cart on a horizontal axis that can move left and right to stabilize the pole vertically up.
The nominal context values for each are summarized in table \ref{tab:nominal_context_values}.

These two systems possess the appealing properties of having smooth dynamics and low dimensions.
As such, they are well suited for studying dimensional analysis in RL.
The control policy is parameterized as a single-layer Radial Basis Function network.
We use Moment Matching \cite{NIPS2002_f3ac63c9} for trajectory predictions as in the original PILCO paper.
For the cartpole, we used the benchmark for distribution shift from \cite{dulacarnold2020realworldrlempirical} and adapted some of the code for our needs.
For the pendulum, we used Gymnasium \cite{towers_gymnasium_2023} on which context variables can be changed with no code modification.

We use two different metrics to evaluate the generalization capabilities of our algorithm.
The return (\ref{eq:return}) is the most commonly used metric used in Markov Decision Processes.
It measures the long-term performance of a controller given an initial state distribution and is computed by a discounted sum of rewards.
The reward for our environments are inversely proportional to the distance from the current state and target 
\begin{equation}
  r_t \propto - d(s_t, s^*), 
  \label{eq:distance_target}
\end{equation}
where $d$ is a distance function in $\mathcal{S}$.
For our specific problems, we only consider finite-time MDPs and thus consider a discount rate $\gamma=1$, which weights identically the rewards from the beginning to end of each episode.

However, during the experiments we realized that this metric was not sufficient to characterise the ability of the controller to stabilize the systems.
The return translates the ability of the agent to stabilize a system at a target position as quickly as possible, which ignores two components of our tasks.
The first is that if two controllers are able to solve the task but one requires more steps to do so, it will be penalized with a lower return since it spends less time in the optimal-rewards regions.
The second inconvenience is that if the controller is able to push the system into a closed-loop equilibrium that deviates from the target, it will not receive an optimal return.
In the next section, we will illustrate these two points for each of the environments we studied.

In order to alleviate the bias of the return metric, we had to find a metric that would translate the ability of the controller to reach a closed-loop equilibrium.
Therefore, we include a binary metric that measures whether in the last step of the episode, the velocity variables of the observations are equal to 0.
We call such an episode \emph{successful}, which allows us to measure the rate of successes for each controller across many different initializations.
Our measure of success rate can be written as
\begin{equation}
\rho = \frac{1}{N} \sum_{i=1}^N  \mathds{1} \left\{\dot{s_T} \leq \epsilon \right\},
\label{eq:success_rates}
\end{equation}
where $N$ is the number of evaluation episodes and $\epsilon$ a threshold.
For our experiments, we used the values $N=100$ and $\epsilon=0.05$.

\subsection{Generalization Score}
\begin{figure}[htb]
  \centering
  \includegraphics[width=0.5\textwidth]{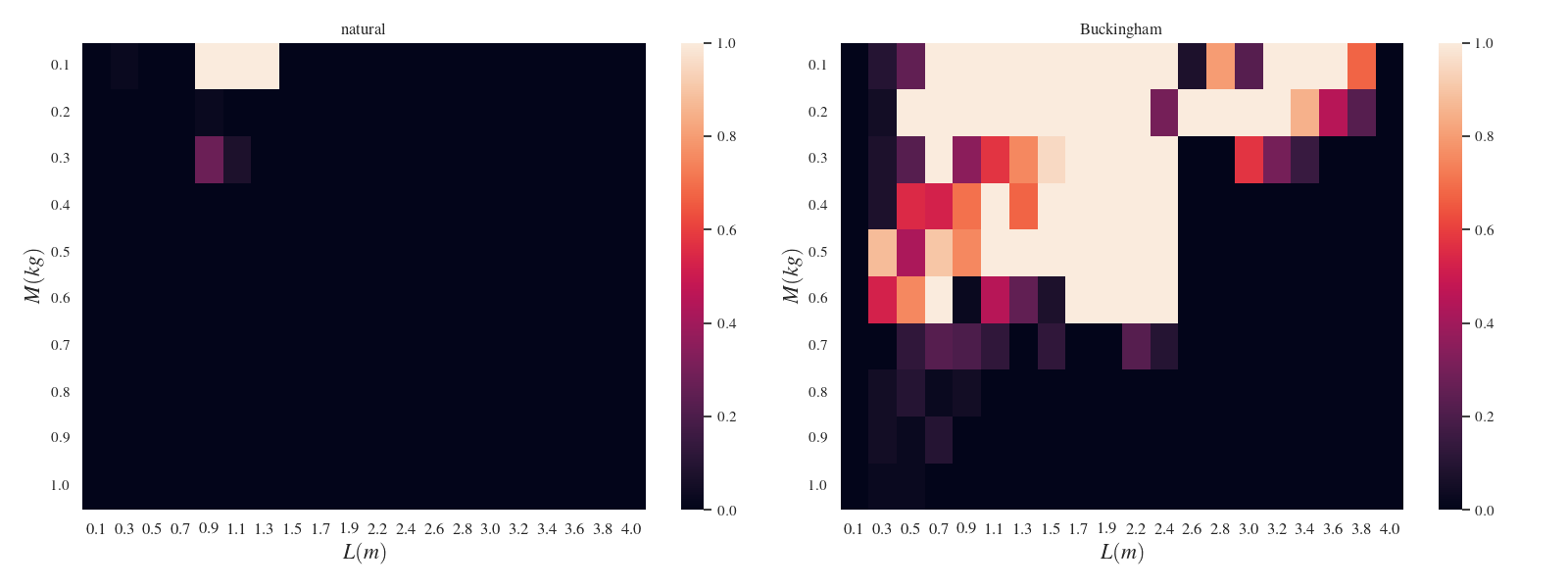}
  \caption[Cart-pole generalization heatmap]{Cartpole success rates when both parameters $L$ and $M$ change simultaneously. We can see how the dimensionless controller (right) can solve the task on a much wider set of context pairs.}
  \label{fig:cartpole-2D_ML-returns}
\end{figure}

We consider the context is the 2-dimensional vector of mass and length. 
We consider perturbations that scale the context vector by a factor $a \in [0.5, 1.5]$ of the nominal context as $\vc = a \vc_0$. 
For the pendulum on figure \ref{fig:pendulum-2D_ML-returns}, both mass $M$ and length $L$ are perturbed around their nominal values.
Bright colours indicate higher return, meaning good generalization of the controller, whereas dark ones indicate failure to stabilize the system.
The first thing we notice is that even in the natural case, the policy for the pendulum is already able to generalize to a significant range of values.
We hypothesize that this is due to the probabilistic nature of the policy search, which presents naturally robust capabilities \cite{charvet2021}.
Nevertheless, the Buckingham transformation is able to enlarge this region, allowing large values of $L$ when is small (below 0.5).
It also allows larger $M$ values, up to 2.4 when $L$ is smaller than 0.9.
The same experiment for the cartpole is presented in figure \ref{fig:cartpole-2D_ML-returns}.

Following the idea of complementing the return metric, we propose another metric that is specific to the problem of generalization and robustness.
We call \emph{controllable area} the surface in parameter space on which the performance of the controller drops by a given fraction $\tau \in [0, 1]$.
The area can be mathematically described as follows,
\begin{equation}
  \mathcal{S}_{control}(\tau) = \left\{\vc \in \gC,~ R_\pi(\vc) \geq \tau R^*(\pi) \right\}.
\label{eq:control_surface_eq}
\end{equation}

This definition allows us to measure the region in context space in which the controller works close to its optimal regime.
We can compute that value with means of an integral over the rate of episodes that have returns greater than the threshold for each infinitesimal context as,
\begin{equation}
  \mathcal{S}_{control}(\tau) = \int_{\mathcal{C}}  \mathds{1}[R_\pi(\vc) \geq \tau R^*(\pi)] d\vc.
\label{eq:control_surface_int}
\end{equation}

\begin{figure}[hbtp]
  \centering
  \begin{subfigure}{.5\textwidth}
    \includegraphics[width=\textwidth]{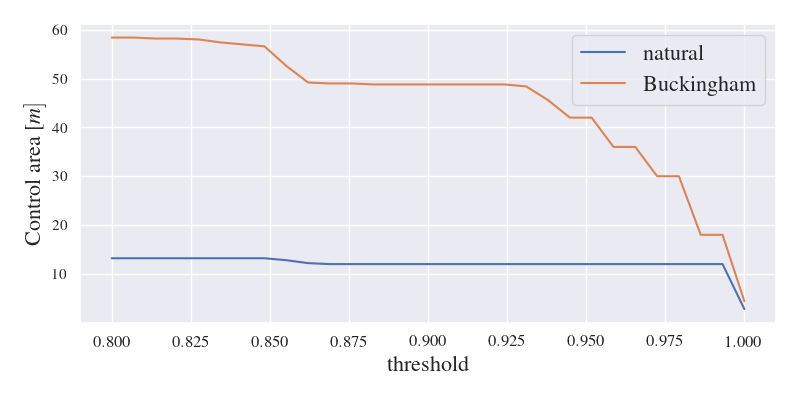}
    \caption{Pole length}
    \end{subfigure}
    \begin{subfigure}{.5\textwidth}
      \includegraphics[width=\textwidth]{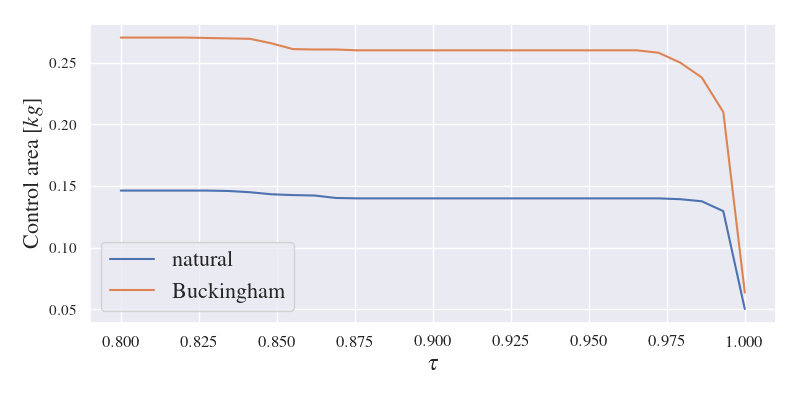}
      \caption{Pole mass}
    \end{subfigure}
    \caption{Control area (\ref{eq:control_surface_eq}) pole length and mass. Higher values on the $x$-axis indicate better generalization.}
    \label{fig:1D_control_area}
\end{figure}

We plot this area as a function of the performance dropoff $\tau$ on figure \ref{fig:1D_control_area} for the pole mass and length.
This figure confirms the findings from above, as we can see the area of optimality of the controllers is much larger for the one in dimensionless space.
Note that depending on the system at hand, the controllable region may not be compact set of the context space.
It is a similar phenomenon that we observe on figure \ref{fig:cartpole-2D_ML-returns}.

\subsection{Action Equivariance}
\begin{figure}[ht]
\centering
\includegraphics[width=.5\textwidth]{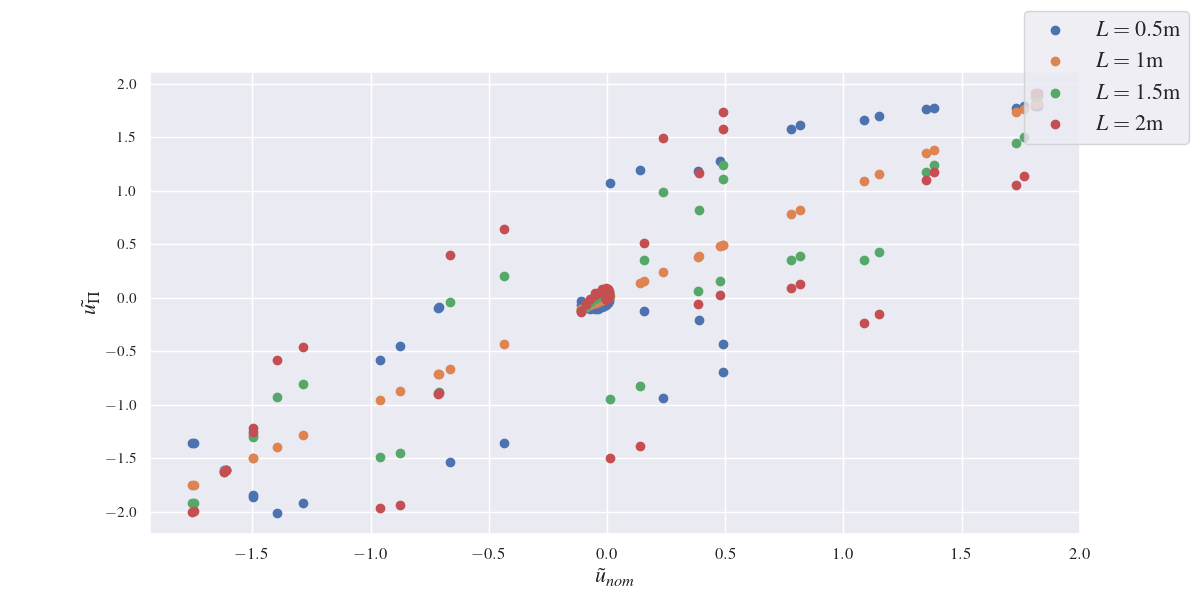}
\caption[Action equivariance]{We plot the actions by the nominal controller ($x$-axis) against the perturbed ones ($y$-axis) for different pole lengths. All the controllers share the same natural input, but it is transformed by the context-dependent $\Pi$-groups.}
\label{fig:actions_equivariance}
\end{figure}
We sample a subset of 100 one-step transitions from the data collected during training.
We then plot the controller actions with the input state going through the Buckingham power-law transformation with appropriate context.
As the environment undergoes a scaling transformation of the context, the dimensionless control actions are also scaled accordingly.
On the other hand, the natural controller is agnostic to the context change and thus not able to stabilize the cartpole and solve the task.
This phenomenon is highlighted on figure \ref{fig:actions_equivariance}.
Here we plot the natural against dimensionless controllers actions on the same sample of state data, but for different contexts.
As we can see, the Buckingham actions are rescaled to reflect the change in pole length.
This ability to transform the controller input further explains how zero-shot generalization can be improved with no additional training data.

\section{Discussion}
\label{sec:discussion}
In this work, we investigated the problem of controller generalization when a dynamic system is subjected to environmental perturbations.
We introduced the dimensionless Markov Decision Process in section \ref{sec:methods}, which allows an autonomous agent to take actions in a dimensionless observation space.
The $\Pi$-MDP is a rescaling of a C-MDP state and action spaces such that each variable becomes dimensionless.
The resulting state-action space stems from additional assumptions about the units of the system and the observation of perturbing variables.
The equivariance properties of the transformation allow \emph{zero-shot transfer} from one context to another.

From the $\Pi$-MDP formulation, we derived a generic framework for model-based policy search that we applied with a Gaussian Process dynamics model (algorithms \ref{alg:interaction-Pi-MDP} and \ref{alg:PS-Pi-MDP}).
The new algorithm we proposed, built on top of PILCO, maintains its data-efficency and improves greatly its generalization capabilities with no further data collection.

We demonstrated empirically that this approach yields controllers that are invariant with respect to the context, provided it can be observed or measured.
Our experiments focused on two different environments, an underactuated pendulum (figure \ref{fig:pendulum-2D_ML-returns}) and a cartpole (figure \ref{fig:cartpole-2D_ML-returns}).
Our results show strong generalization properties of the controller when the physical properties of the system, such as pole length and mass, drift from their initial training value.
While these are simple systems, because of their second-order dynamics and low dimension, the consistency of the results suggests the methodology could be successfully applied to more complex systems, which we leave to future work.

Conceptually, our approach comes within the scope of instilling physics prior knowledge in Machine Learning pipelines to increase model robustness \cite{NEURIPS_DATASETS_AND_BENCHMARKS2021_f033ab37}.
The main weakness of this approach is the requirement for measuring what the perturbation variables are at any point in the deployment of the controller.
We believe identification of the parameters could be achieved with different control policies that aim to actively infer those values based on exploratory trajectories and leave this direction for future work.
The second limitation of this approach is the requirement for knowing the measurements' dimensions which can be prohibitively expensive on high-dimensional systems.
To alleviate this, one could either use physical priors to determine which transformation is most suited to type of perturbation that might be later encountered or find them numerically as in \cite{bakarji2022dimensionally}.

\FloatBarrier
\pagebreak

\appendix\section{Pendulum $\Pi$-groups}
\label{app:pendulum_pi_groups}
Dynamic variables
\begin{equation}
\begin{cases}
  u: & \begin{bmatrix}1 & 1 & -2 \end{bmatrix} \\    
  \theta:& \begin{bmatrix}0 & 0 & 0 \end{bmatrix} \\
  \dot{\theta}:& \begin{bmatrix}0 & 0 & -1 \end{bmatrix} \\
  \ddot{\theta}:& \begin{bmatrix}0 & 0 & -2 \end{bmatrix} \\
\end{cases}
\end{equation}

\noindent Context
\begin{equation}
\begin{cases}
  M: & \begin{bmatrix}1 & 0 & 0 \end{bmatrix} \\
  g: & \begin{bmatrix}0 & 1 & -2 \end{bmatrix} \\
  L: & \begin{bmatrix}0 & 1& 0 \end{bmatrix}\\
\end{cases}
\end{equation}

\noindent The context matrix
\begin{equation}
\mC  = \begin{bmatrix}
1 & 0 & 0 \\
0 & 1 & -2 \\
0 & 1 & 0 \end{bmatrix}
\end{equation}
is full rank and thus the variables $(M,g,L)$ can be used for non-dimensionalizing the other ones.

\begin{equation}
\begin{cases}
  \left[u^{\alpha_u} \cdot M^{\beta_u} \cdot g^{\delta_u} \cdot L^{\gamma_u}  \right] = 0 \\
  \left[\theta^{\alpha_\theta} \cdot M^{\beta_\theta} \cdot g^{\delta_\theta} \cdot L^{\gamma_\theta}  \right] = 0 \\
  \left[\dot{\theta}^{\alpha_{\dot{\theta}}} \cdot M^{\beta_{\dot{\theta}}} \cdot g^{\delta_{\dot{\theta}}} \cdot L^{\gamma_{\dot{\theta}}}  \right] = 0 \\
  \left[\ddot{\theta}^{\alpha_{\ddot{\theta}}} \cdot M^{\beta_{\ddot{\theta}}} \cdot g^{\delta_{\ddot{\theta}}} \cdot L^{\gamma_{\ddot{\theta}}}  \right] = 0 \\
\end{cases}
\label{eq:system_coefs}
\end{equation}
Where the bracket signs $\left[x\right]$ represent the dimension of variable $x$ and each power law within equation \ref{eq:system_coefs} will be the $\Pi$-groups.

Because we know the dimension of the variables $u, \theta, \dot\theta, \ddot\theta$ and because $[x\times y]=[x]\times [y]$ the system can be rewritten as

\begin{equation}
\begin{cases}
  M^{\beta_u} \cdot L^{\alpha_u+\delta_u + \gamma_u} \cdot t^{-2\delta_u + \alpha_u} = 1 \\
  M^{\beta_{\dot\theta}} \cdot L^{\delta_{\dot\theta} + \gamma_{\dot\theta}} \cdot t^{-2 \delta_{\dot\theta} -1} = 1  \\
  M^{\beta_{\ddot\theta}} \cdot L^{\delta_{\ddot\theta} + \gamma_{\ddot\theta}} \cdot t^{-2 \delta_{\dot\theta} -2} = 1
\end{cases}
\label{eq:system_to_solve}
\end{equation}
We removed the equation for $\theta$ because as an angle, this variable is naturally dimensionless.
The coefficients are found by solving one system for each variable.

\subsection{Torque $u$}
$\Pi_u= u^{\alpha} \cdot M^{\beta} \cdot g^{\delta} \cdot L^{\gamma}$
Using the first term from \ref{eq:system_to_solve} and replacing the terms by their dimension we obtain,
\begin{equation}
  M^{\alpha+\beta}. L^{\alpha + \delta + \gamma} t^{-2\alpha -2\delta} = 1.
\end{equation}

All exponents must be 0 to ensure the homogeneity which yields
\begin{equation}
  \begin{cases}
    \alpha +\beta & = 0 \\
    \alpha + \delta + \gamma & = 0\\
    \alpha - 2\delta & = 0
  \end{cases}
\end{equation}

The last equation implies $\alpha + \delta = 0$, which we substract from the first equation to obtain
\begin{equation}
\begin{cases}
  \beta = \delta\\
  \alpha + \beta = 0\\
  \alpha + \delta + \gamma = 0,
\end{cases}
\end{equation}
and then using $\alpha +  \delta = 0$
\begin{equation}
\begin{cases}
  \beta = \delta
  \gamma = 0 \\
  \alpha + \beta = 0. \\
\end{cases}
\end{equation}
Because the solution is not unique, we choose $\alpha=1$, which gives the dimensionless torque
\begin{equation}
  \Pi_u = \frac{u}{Mg}.
\end{equation}

\subsection{Angular speed $\dot\theta$}
$\Pi_{\dot{\theta}}= \dot{\theta}^{\alpha} \cdot M^{\beta} \cdot g^{\delta} \cdot L^{\gamma}$
We replace the variables with their dimensions to obtain
\begin{equation}
M^{\beta}. L^{\delta + \gamma} . t^{-\alpha - 2\delta} = 1,
\end{equation}
which we can solve with the systems
\begin{equation}
\begin{cases}
  \beta = 0 \\
  \delta + \gamma = 0\\
  \alpha + 2 \delta = 0.
\end{cases}
\end{equation}
By subtracting twice the second equation from the third we obtain
\begin{equation}
\begin{cases}
  \beta = 0 \\
  \alpha = 2 \gamma \\
  \delta + \gamma = 0.
\end{cases}
\end{equation}
We choose $\delta=1$ yielding
\begin{equation}
  \Pi_{\dot{\theta}} = \dot{\theta}^2 \dfrac{g}{L}.
\end{equation}

\subsection{Angular acceleration $\ddot\theta$}
By the same process we obtain,
\begin{equation}
  M^{\beta}. L^{\delta + \gamma} . t^{-\alpha - 2\delta} = 1,
\end{equation}
$\beta=0$, so we have the systems
\begin{equation}
\begin{cases}
  \delta + \gamma = 0 \\
  \alpha = \beta.
\end{cases}
\end{equation}
This yields
\begin{equation}
  \Pi_{\ddot{\theta}} = \ddot{\theta} \frac{g}{L}.
\end{equation}


\section{Cartpole $\Pi$-groups}
\label{app:cartpole_pi_groups}
The movement of the cartpole depends on the variables $(x, \cos(\theta), \sin(\theta), \dot{x}, \dot{\theta}), u$.
A trivial $\Pi$-group for the cart position is $\Pi_x = \dfrac{x}{L}$, where $L$ is the pole length.
For the angular speed, we use the same transformation as the pendulum.
Therefore, we  need to compute the dimensionless variables for $\dot{x}$ and $u$

\subsection{Cart speed $\dot{x}$}
With $\Pi_{\dot{x}} = \dot{x}^\alpha \cdot M^{\beta} \cdot g^{\delta} \cdot L^{\gamma}$,
we obtain with $[\dot{x}]=L.t^{-1}$,
\begin{equation}
  M^\beta . L^{\alpha + \delta + \gamma} . t^{-\alpha - 2\delta}
\end{equation}
which yields $\beta=0$.
We then substract on equation with the other to obtain,
\begin{equation}
\begin{cases}
  \delta - \gamma = 0\\
  \alpha + 2\delta = 0,
\end{cases}
\end{equation}
which is solved with $\delta=\gamma=-1$.
Therefore the dimensionless variable for the cart is
\begin{equation}
  \Pi_{\dot{x}} = \dfrac{\dot{x}^2}{Lg}.
\end{equation}

\subsection{Force $u$}
$\Pi_u= u^{\alpha} \cdot M^{\beta} \cdot g^{\delta} \cdot L^{\gamma}$
The dimension of the control force is $[u]=M.L.t^{-2}$.
Using that value yields the system
\begin{equation}
\begin{cases}
  \alpha + \beta = 0 \\
  \alpha + \delta = 0 \\
  \alpha + \delta + \gamma = 0
\end{cases}
\end{equation}
and by substracting the first two equations we obtain
\begin{equation}
\begin{cases}
  \gamma = 0\\
  \beta = \delta\\
  \alpha + \delta = 0.
\end{cases}
\end{equation}
With $\alpha=1$, we obtain the resulting
\begin{equation}
  \Pi_u = \dfrac{u}{Mg}.
\end{equation}


\label{sec:reference_examples}

\bibliography{aaai25}

\end{document}